\documentclass[runningheads]{llncs}
\usepackage{graphicx}
\usepackage{url}
\usepackage[hidelinks]{hyperref}
\usepackage{booktabs}
\begin{document}
\title{FoundationLayerNorm: Scaling BERT and GPT to 1,000 Layers}
%
%
\author{Dezhou Shen}
%
\authorrunning{D. Shen}
%
\institute{
\email{sdz15@tsinghua.org.cn}
}
\maketitle              
\begin{abstract}
The mainstream BERT/GPT model contains only 10 to 20 layers, and there is little literature to discuss the training of deep BERT/GPT.
This paper proposes a simple yet effective method to stabilize BERT and GPT training. We successfully scale up BERT and GPT to 1,000 layers, which is an order of magnitude deeper than previous BERT and GPT.
The proposed method FoundationLayerNormalization enables efficient training of deep neural networks and is validated at the 1000-layer scale.

\end{abstract}

\section{Introduction}

In recent years, there has been a trend in large Transformer models, with capacities increasing dramatically from millions of parameters to billions, or even trillions.
Large models yield state-of-the-art performance on a variety of tasks and demonstrate impressive capabilities in few-shot and zero-shot learning.
However, scaling to 1,000 layers for BERT or GPT has rarely been reported.
To improve the training stability of the BERT and GPT with 1,000 layers, we introduce a new normalization function (FoundationLayerNorm) at the residual connections with a theoretical justification for model updating via constant constraints.
This approach improves the stability of the BERT and GPT, so we are able to scale the model depth to more than 1,000 layers.
The proposed method can be a preferred alternative for transformers, not only for extremely deep models, but also for existing large models.

\par In summary, the contributions of this article are as follows:
\begin{itemize}
    \item We proposed a Layer Normalization method, which was used to train a BERT~\cite{devlin2019bert} model with 1,000 layers, which to the best of our knowledge, is the deepest BERT model.
    \item We proposed a Layer Normalization method, which was used to train a GPT~\cite{radford2018improving} model with 1,000 layers, which to the best of our knowledge, is the deepest GPT model.
\end{itemize}


\section{Background and Related Work}
\par One of the challenges of deep learning is that the gradients with respect to the weights in a layer are highly dependent on the outputs of neurons in the previous layer, especially if these outputs vary in a highly correlated way. The Batch normalization~\cite{ioffe2015batch} method normalizes the summed input of each hidden unit on the training instance. Specifically, for the input in the ith layer, the batch normalization method resizes the input according to the variance under the data distribution, according to the sum of the data.
\par Liu et. al~\cite{liu2020understanding} showed that the decoder is more unstable than the encoder.
\par Ba et. al~\cite{ba2016layer} found that one way to reduce training time is to regularize the activity of neurons by computing the normalized mean and variance used to sum the inputs to neurons in a layer from all the summed inputs in a single training situation , also giving each neuron its own adaptive bias and gain, which are applied after normalization and before nonlinearity; layer normalization is very effective for stabilizing hidden state dynamics in recurrent networks.
\par Nguyen et. al~\cite{nguyen2019transformers} found that the pre-norm residual connection (Pre-LN) improves the stability of the transformer based on the post-norm connection (post-LN). However, the gradient of the front LN at the bottom layer tends to be larger than the top layer~\cite{shleifer2021normformer}, resulting in a performance degradation compared to the post LN.
\par Wang et. al~\cite{wang2022deepnet} proposed the DeepNorm normalization method, extending the transformer to 1,000 layers, the authors scaled the residual connections before performing layer normalization, only the weights of the feedforward network, and the value projection and output projection of the attention layer were scaled.

\section{Deep Layer Normalization}

The formulation of DEEPNORM~\cite{wang2022deepnet} can be written as:
 $ x_{i+1}=LN(\alpha x_i+G_i(x_i, \theta_i)) $
where $\alpha$ is a constant and G is a function of the i-th Transformer sublayer (i.e. attention or feedforward network) with parameter $\theta$.
Furthermore, DEEPNORM scales the weights $\theta$ within the residual branch by $\beta$.
Given a BERT model of N-layer enoder, $ \alpha=(2N)^{1/4} $.
Given the transformer of N-layer encoder, M-layer decoder, for Transformer-Encoder $ \alpha=0.81(N^4M)^{1 /16} $ and for Transformer-Decoder $ \alpha=(3N)^{1/4} $.

\section{Scaling BERT to 1,000 Layers}
We processed 9G data as training data, and filtered out code documents from The Pile English dataset.

\subsection{Upscale Layer Normalization}

We revise the DEEPNORM by keeping the Layer Normalization but holding the weights $\theta$ within the residual branch, instead of scaling with $\beta$ as proposed in DEEPNORM by Wang et. al.
\par The Upscale Layer Normalization formula can be written as:
 $ x_{i+1}=LN(\alpha x_i+G_i(x_i, \theta_i)) $
where $\alpha$ is a constant and G is a function of the i-th BERT sublayer (i.e. attention or feedforward network) with parameter $\theta$.
Given a BERT model of N-layer enoder, $ \alpha=(2N)^{1/4} $.

We use an Nvidia 3090 GPU to pretrain the BERT model for 100k steps for four days, and the loss is 39.6 and the perplex is 4.85e8.

\subsection{BERT Model Architecture}
We use the original BERT~\cite{devlin2019bert} tokenizer for our training, which has a vocabulary of 30522.
The network structure and hyper-parameters of the model are as follows in Table-\ref{tab:bertparam}:

\begin{table}[]
       \begin{center}
\begin{tabular}{ll}
        \toprule
\textbf{parameters} & \textbf{value} \\
    \midrule
num-parameters      & 52M            \\
num-layers          & 1,000           \\
hidden-size         & 64             \\
num-attention-heads & 2              \\
seq-length          & 512            \\
learning rate       & 1e-4           \\
min-learning rate   & 1e-5           \\
lr-decay-style      & linear         \\
lr-warmup-fraction  & 0.01           \\
weight-decay        & 0.01           \\
fp16                & True          \\
    \bottomrule
\end{tabular}
    \caption{Hyper-parameters for 1,000 layer BERT model.}
        \label{tab:bertparam}
       \end{center}
\end{table}

\section{Foundation Layer Normalization}
The formulation of Foundation Layer Normalization can be written as:
$ x_{i+1}=LN(0.974 x_i+G_i(x_i, \theta_i)) $
where $0.974$ is a constant parameter from experience and G is a function of the i-th GPT sublayer (i.e. attention or feedforward network) with parameter $\theta$.

We processed 200G data as training data, and filtered out code documents from The Pile English dataset.
We use an Nvidia 3090 GPU to pretrain the GPT model for 150k steps for seven days, and the loss is 1.28 and the perplex is 3.6.

We use the original GPT-2~\cite{radford2019language} tokenizer for our training, which has a vocabulary of 50257.
The GPT Model network structure and hyper-parameters of the model are as follows in Table-\ref{tab:gptparam}:

\begin{table}[]
       \begin{center}
\begin{tabular}{ll}
        \toprule
\textbf{parameters} & \textbf{value} \\
    \midrule
num-parameters      & 815.5M           \\
num-layers          & 1,000           \\
hidden-size         & 256            \\
num-attention-heads & 1              \\
seq-length          & 1024           \\
optimizer           & Adam           \\
learning rate       & 1e-5           \\
lr-decay-style      & cosine         \\
lr-warmup-fraction  & 0.01           \\
weight-decay        & 0              \\
fp32                & True           \\
    \bottomrule
\end{tabular}
    \caption{Hyperparameters for 1,000 layer GPT model.}
        \label{tab:gptparam}
       \end{center}
\end{table}

\section{Experiments}

\subsection{BERT QQP evaluation}
Quora Question Pairs (QQP) is a social QA question task proposed by Wang~\cite{wang2018glue}. QQP belongs to a similarity and paraphrase task.
There are 364k in the training set and 391k in the test set. We use an Nvidia-3090 GPU to evaluate the BERT model defined in Table-\ref{tab:bertparam},
and we evaluated for one epoch with batch size of 16, learning rate of 1e-5. The performance result is shown in Table-\ref{tab:bertqqptest}.

\begin{table}[]
       \begin{center}
\begin{tabular}{lllll}
        \toprule
\textbf{task} & \textbf{precision}& \textbf{recall}& \textbf{f1-score}& \textbf{accuracy} \\
    \midrule
QQP-test      &  72\% &69\%  & 70\%  & 73\%          \\
    \bottomrule
\end{tabular}
    \caption{QQP evaluation for 1,000 layer BERT model.}
        \label{tab:bertqqptest}
       \end{center}
\end{table}

\subsection{GPT QQP evaluation}
We evaluate the GPT on Quora Question Pairs~\cite{wang2018glue}, LAMBDA~\cite{paperno2016lambada}, WinoGrande~\cite{sakaguchi2020winogrande}, Hellaswag~\cite{zellers2019hellaswag} and PIQA~\cite{bisk2020piqa}.
We use an Nvidia-3090 GPU to evaluate the GPT model defined in Table-\ref{tab:gptparam}.
\par LAMBADA is a collection of narrative passages sharing the characteristic that human subjects are able to guess their last word if they are exposed to the whole passage, but not if they only see the last sentence preceding the target word.
WinoGrande is a collection of 44k problems, inspired by Winograd Schema Challenge (Levesque, Davis, and Morgenstern 2011), but adjusted to improve the scale and robustness against the dataset-specific bias.
Formulated as a fill-in-a-blank task with binary options, the goal is to choose the right option for a given sentence which requires commonsense reasoning.
Hellaswag is a commonsense inference challenge dataset.
Physical Interaction: Question Answering (PIQA) is a physical commonsense reasoning and a corresponding benchmark dataset.
The performance result is shown in Table-\ref{tab:gpttest}.

\begin{table}[]
       \begin{center}
\begin{tabular}{ll}
        \toprule
\textbf{task} & \textbf{score} \\
    \midrule
QQP-Acc      &  46.93\%          \\
QQP-F1       &  48.37\%          \\
LAMBADA-PPL  &  8.37E5           \\
LAMBADA-Acc  &  0.72\%           \\
Winogrande-Acc  &  50.36\%       \\
Hellaswag-Acc & 25.54\%          \\
PIQA-Acc  &     55.17\%          \\
    \bottomrule
\end{tabular}
    \caption{QQP evaluation for 1,000 layer GPT model.}
        \label{tab:gpttest}
       \end{center}
\end{table}

\subsection{Discussion}

Seen from Table-\ref{tab:gptcomparison}, with 1/75 parameter size of GPT-J, the GPT-1k achieved competing performance in PIQA and Winogrande datasets.

\begin{table}[]
       \begin{center}
\begin{tabular}{llllll}
        \toprule
\textbf{Model}    & \textbf{FLOPs} & \textbf{LAMBADA} & \textbf{Winogrande} & \textbf{Hellaswag} & \textbf{PIQA}   \\
    \midrule
GPT-2 1.5B        & -                       & 51.21\%          & 59.4\%              & 50.9\%             & 70.8\%          \\
GPT-Neo 1.3B      & 3.0e21                  & 57.2\%           & 55.0\%              & 48.9\%             & 71.1\%          \\
GPT-Neo 2.7B      & 6.8e21                  & 62.2\%           & 56.5\%              & 55.8\%             & 73.0\%          \\
\textbf{GPT-J 6B\cite{gpt-j}} & \textbf{1.5e22}         & \textbf{69.7\%}  & \textbf{65.3\%}     & \textbf{66.1\%}    & \textbf{76.5\%} \\
GPT-1k 815.5M(ours)     & 3.72e19                 & 0.72\%           & 50.36\%             & 25.54\%            & 55.17\%          \\
    \bottomrule
\end{tabular}
    \caption{GPT-1k model performance comparison with accuracy of LAMBADA, Winogrande, Hellaswag and PIQA datasets are compared with baseline models.}
        \label{tab:gptcomparison}
       \end{center}
\end{table}

\section{Conclusion}

In this paper, two Layer Normalization methods, named Upscale Layer Normalization and Foundation Layer Normalization, are proposed,
and the effectiveness of the methods is verified in the BERT and GPT network structures.
The proposed method LayerNorm enables efficient training of deep neural networks and is validated at the 1000-layer scale.
The results indicate that depth is a promising extension direction.

\section{Future Work}

With the development of hardware and software, we will try more efficient model tricks for the deeper network layers and train models based on BERT and GPT in the future.

\bibliographystyle{splncs04}
\bibliography{founder-ln-arxiv}

\begin{thebibliography}{10}

\bibitem{ba2016layer}
Jimmy~Lei Ba, Jamie~Ryan Kiros, and Geoffrey~E Hinton.
\newblock Layer normalization.
\newblock {\em arXiv preprint arXiv:1607.06450}, 2016.

\bibitem{bisk2020piqa}
Yonatan Bisk, Rowan Zellers, Jianfeng Gao, Yejin Choi, et~al.
\newblock Piqa: Reasoning about physical commonsense in natural language.
\newblock In {\em Proceedings of the AAAI conference on artificial
  intelligence}, volume~34, pages 7432--7439, 2020.

\bibitem{devlin2019bert}
Jacob Devlin, Ming-Wei Chang, Kenton Lee, and Kristina Toutanova.
\newblock {BERT}: Pre-training of deep bidirectional transformers for language
  understanding.
\newblock In {\em Proceedings of the 2019 Conference of the North {A}merican
  Chapter of the Association for Computational Linguistics: Human Language
  Technologies, Volume 1 (Long and Short Papers)}, pages 4171--4186,
  Minneapolis, Minnesota, 2019. Association for Computational Linguistics.

\bibitem{ioffe2015batch}
Sergey Ioffe and Christian Szegedy.
\newblock Batch normalization: Accelerating deep network training by reducing
  internal covariate shift.
\newblock In {\em International conference on machine learning}, pages
  448--456. PMLR, 2015.

\bibitem{liu2020understanding}
Liyuan Liu, Xiaodong Liu, Jianfeng Gao, Weizhu Chen, and Jiawei Han.
\newblock Understanding the difficulty of training transformers.
\newblock In {\em Proceedings of the 2020 Conference on Empirical Methods in
  Natural Language Processing (EMNLP)}, pages 5747--5763, 2020.

\bibitem{nguyen2019transformers}
Toan~Q Nguyen and Julian Salazar.
\newblock Transformers without tears: Improving the normalization of
  self-attention.
\newblock {\em arXiv preprint arXiv:1910.05895}, 2019.

\bibitem{paperno2016lambada}
Denis Paperno, Germ{\'a}n Kruszewski, Angeliki Lazaridou, Quan~Ngoc Pham,
  Raffaella Bernardi, Sandro Pezzelle, Marco Baroni, Gemma Boleda, and Raquel
  Fern{\'a}ndez.
\newblock The lambada dataset: Word prediction requiring a broad discourse
  context.
\newblock {\em arXiv preprint arXiv:1606.06031}, 2016.

\bibitem{radford2018improving}
Alec Radford, Karthik Narasimhan, Tim Salimans, and Ilya Sutskever.
\newblock Improving language understanding by generative pre-training.
\newblock {\em OpenAI blog}, 2018.

\bibitem{radford2019language}
Alec Radford, Jeffrey Wu, Rewon Child, David Luan, Dario Amodei, Ilya
  Sutskever, et~al.
\newblock Language models are unsupervised multitask learners.
\newblock {\em OpenAI blog}, 1(8):9, 2019.

\bibitem{sakaguchi2020winogrande}
Keisuke Sakaguchi, Ronan Le~Bras, Chandra Bhagavatula, and Yejin Choi.
\newblock Winogrande: An adversarial winograd schema challenge at scale.
\newblock In {\em Proceedings of the AAAI Conference on Artificial
  Intelligence}, volume~34, pages 8732--8740, 2020.

\bibitem{shleifer2021normformer}
Sam Shleifer, Jason Weston, and Myle Ott.
\newblock Normformer: Improved transformer pretraining with extra
  normalization.
\newblock {\em arXiv preprint arXiv:2110.09456}, 2021.

\bibitem{wang2018glue}
Alex Wang, Amanpreet Singh, Julian Michael, Felix Hill, Omer Levy, and Samuel
  Bowman.
\newblock Glue: A multi-task benchmark and analysis platform for natural
  language understanding.
\newblock In {\em Proceedings of the 2018 EMNLP Workshop BlackboxNLP: Analyzing
  and Interpreting Neural Networks for NLP}, pages 353--355, 2018.

\bibitem{gpt-j}
Ben Wang and Aran Komatsuzaki.
\newblock {GPT-J-6B: A 6 Billion Parameter Autoregressive Language Model}.
\newblock \url{https://github.com/kingoflolz/mesh-transformer-jax}, May 2021.

\bibitem{wang2022deepnet}
Hongyu Wang, Shuming Ma, Li~Dong, Shaohan Huang, Dongdong Zhang, and Furu Wei.
\newblock Deepnet: Scaling transformers to 1,000 layers.
\newblock {\em arXiv preprint arXiv:2203.00555}, 2022.

\bibitem{zellers2019hellaswag}
Rowan Zellers, Ari Holtzman, Yonatan Bisk, Ali Farhadi, and Yejin Choi.
\newblock Hellaswag: Can a machine really finish your sentence?
\newblock {\em arXiv preprint arXiv:1905.07830}, 2019.

\end{thebibliography}
\end{document}